\documentclass[letterpaper, 10 pt, conference]{ieeeconf}  % Comment this line out if you need a4paper

\IEEEoverridecommandlockouts                              % This command is only needed if 
\usepackage{graphics} % for pdf, bitmapped graphics files
\usepackage{epsfig} % for postscript graphics files
\usepackage{mathptmx} % assumes new font selection scheme installed
\usepackage{times} % assumes new font selection scheme installed
\usepackage{amsmath} % assumes amsmath package installed
\usepackage{amssymb}  % assumes amsmath package installed
\usepackage{graphicx}
\usepackage[table]{xcolor} 
\usepackage{cite,comment}
\usepackage{caption}    %For images and and smaller images
\usepackage{subcaption}
\usepackage{multirow}
\usepackage{soul}
\usepackage{booktabs}
\usepackage[nolist]{acronym} % acronyms by the \ac{label} command

\definecolor{orange}{RGB}{252, 130, 62}
\definecolor{red}{RGB}{255, 0, 0}
\definecolor{brown}{RGB}{155, 25, 10}
\definecolor{blue}{RGB}{0, 0,255}
\definecolor{green}{RGB}{78, 196, 164}
\definecolor{burgundy}{RGB}{127, 0, 33}

\newcommand{\etal}{\textit{et al}.}

\newacro{ec}[EC]{Elastic Context}
\newacro{gnn}[GNN]{Graph Neural Network}
\newacro{pbd}[PBD]{Position-based Dynamic}
\newacro{mlp}[MLP]{Multi-layer Perceptron}

 \title{\LARGE \bf
 Elastic Context: Encoding Elasticity for Data-driven Models of Textiles
 }

\author{Alberta Longhini${^1}$, Marco Moletta${^1}$, Alfredo Reichlin${^1}$, Michael C. Welle${^1}$, Alexander Kravberg${^1}$, \\ Yufei Wang${^2}$, David Held${^2}$, Zackory Erickson${^2}$, and Danica Kragic${^1}$% <-this % stops a space
\thanks{$^{1}$The authors are with the Robotics, Perception and Learning Lab, EECS, at KTH Royal Institute of Technology, Stockholm, Sweden
     {\tt\small albertal, moletta, alfrei, mwelle, okr, dani@kth.se}}%
        \thanks{    $^{2}$The authors are with Carnegie Mellon University, Pittsburgh, USA
        {\tt\small yufeiw2, dheld, zerickso@andrew.cmu.edu}}%
}

\begin{document}

\maketitle
\thispagestyle{empty}
\pagestyle{empty}

%%%%%%%%%%%%%%%%%%%%%%%%%%%%%%%%%%%%%%%%%%%%%%%%%%%%%%%%%%%%%%%%%%%%%%%%%%%%%%%%
\begin{abstract}

Physical interaction with textiles, such as assistive dressing or household tasks, requires advanced dexterous skills. The complexity of textile behavior during stretching and pulling is influenced by the material properties of the yarn and by the textile's construction technique, which are often unknown in real-world settings. Moreover, identification of physical properties of textiles through sensing commonly available on robotic platforms remains an open problem. To address this, we introduce Elastic Context (EC), a method to encode the elasticity of textiles using stress-strain curves adapted from textile engineering for robotic applications. We employ EC to learn generalized elastic behaviors of textiles and examine the effect of EC dimension on accurate force modeling of real-world non-linear elastic behaviors.

%Physical interaction with textiles, such as assistive dressing scenarios, relies on advanced dexterous capabilities. The underlying complexity in textile behavior when being pulled and stretched is due to both the yarn material properties and the textile construction technique. Those are often not known a-priori, making it almost impossible to identify through sensing commonly available on robotic platforms. We introduce Elastic Context (EC) to encode elastic behaviors, which enables a more effective physical interaction with textiles.  The definition of EC relies on stress/strain curves commonly used in textile engineering, which we reformulated for robotic applications. We employ EC using a GNN to learn generalized elastic behaviors of textiles. Furthermore, we explore the effect of the dimension of the EC on accurate force modeling of non-linear real-world elastic behaviors. 

\end{abstract}

%Today, there are no commonly adopted and annotated datasets on which the various interaction or property identification methods are assessed. 
%One important property that affects the interaction is material elasticity that results from both the yarn material and construction technique: these two are intertwined and, if not known a-priori, almost impossible to 
%, highlighting the challenges of current robotic setups to sense textile properties. %

%\newpage
%\newpage

\section{Introduction}

% \begin{comment}
% % Motivation paragraph
% Motivated by recent developments in artificial perception, cognition and control, robotic systems have found widespread use in multiple scenarios such as in medical~\cite{peters2018review}, educative~\cite{belpaeme2018social} and industrial settings~\cite{ji2019industrial}. In particular, these developments have allowed the use of robots as suitable tools for assisting humans in complex tasks such as industrial collaboration~\cite{vysocky2016human} and assistive dressing~\cite{klee2015personalized,kapusta2019personalized}. In order to be effective in such tasks, robots require complex controllers that allow the precise manipulation of highly-deformable objects, such as clothes and textiles, whose dynamics can be shaped by the material of the yarns and the way they are arranged to construct fabrics~\cite{longhini2021textile}. 
% \end{comment}

% Motivation paragraph
Manipulation of  deformable objects such as textiles is common in medical robotics~\cite{peters2018review}, human-robot interaction~\cite{erickson2018deep}, automation of household tasks~\cite{verleysen2020video}, assistive dressing~\cite{klee2015personalized,kapusta2019personalized}. However, as discussed in~\cite{chi2022iterative}, textile objects are challenging when it comes to manipulation due to complex dynamics and often unknown physical properties (e.g. elasticity, friction, density distribution). These properties depend on the yarn material and textile construction technique and may be difficult to estimate in online robotics manipulation scenarios~\cite{ha2022flingbot}.

\begin{figure}[t]
  \centering
   \includegraphics[width=1.\linewidth]{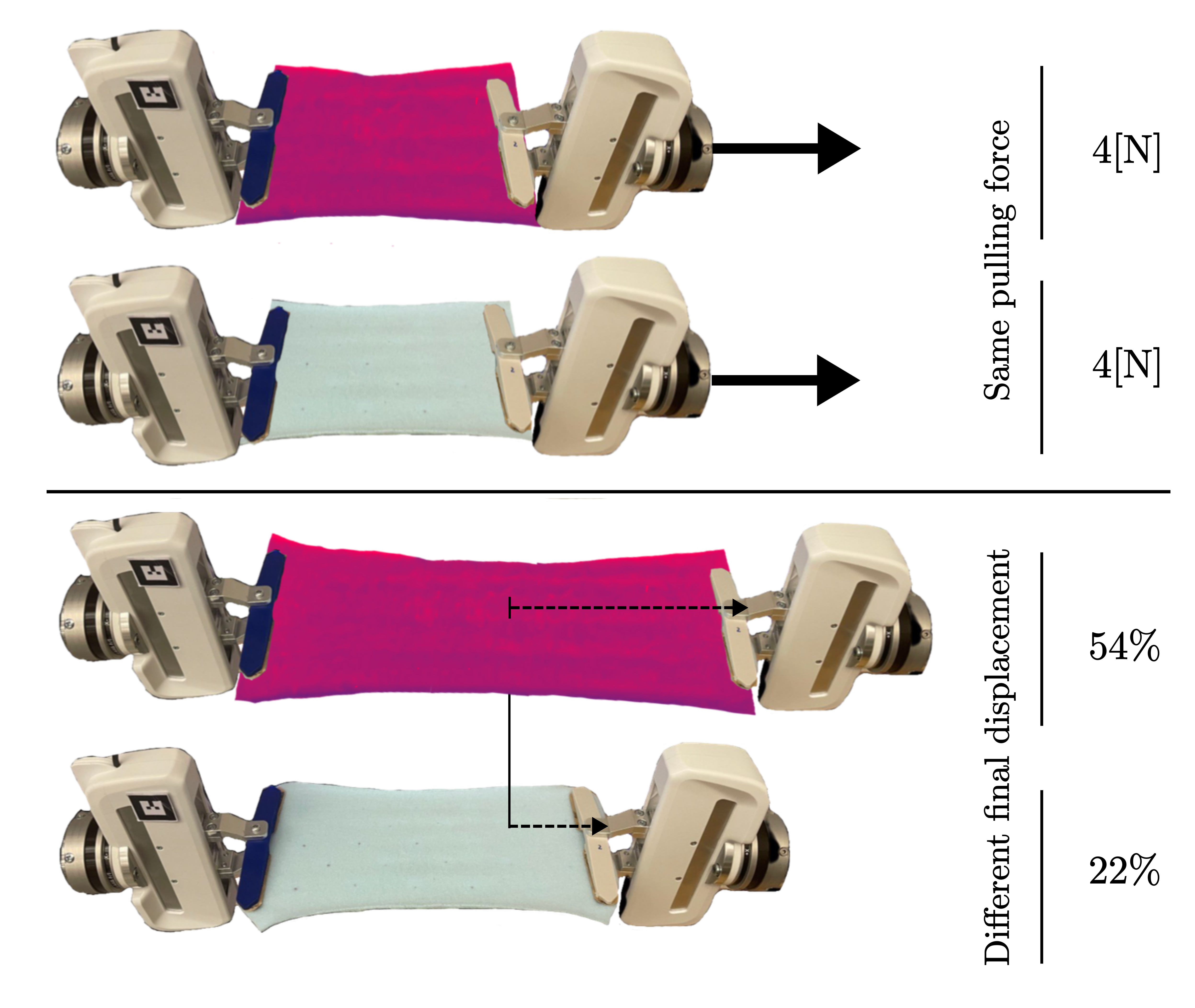}
  
  \caption{The role of elasticity in the manipulation of different textile samples, exhibiting different behaviors under the same pulling force.}
  \label{fig:different_meshes}
  \vspace{-\baselineskip}
\end{figure}

% Zoom-in on the problem paragraph
We address the problem of encoding elastic properties of textile objects, to learn data-driven models that generalize to variations of elastic properties, see  Fig.~\ref{fig:different_meshes}. Modeling the elastic behavior of textiles has been addressed in two rather distant communities: textile engineering~\cite{poincloux2018geometry} and computer graphics~\cite{clyde2017modeling, sperl2020homogenized}.
%Modeling the elastic behavior of deformable objects has previously been explored using visual information~\cite{bhat2003estimating, duan2021learning}. Robotic manipulation tasks, on the other hand, require an  understanding of their intrinsic physical properties, such as elasticity~\cite{duenser2018interactive} and textile construction techniques~\cite{longhini2021textile}. Analytic models of these properties have been developed in two rather different communities: textile engineering~\cite{poincloux2018geometry} and computer graphics~\cite{clyde2017modeling, sperl2020homogenized}. 
The analytical models from these communities are computationally expensive and commonly not applicable in real-time robotic manipulation. Moreover, such models build on parameters measured with high-precision devices under controlled experiments~\cite{eberhardt1996fast, yousef2018investigating}. Today, there are no commonly adopted and annotated datasets for which these parameters are provided, requiring thus alternative strategies to encode elastic properties  of textiles.

%Recent work has shown how force-torque measurements and simple pulling and twisting robot actions may be used to identify yarn materials and construction techniques, making a step towards robotic understanding of deformable objects properties~\cite{longhini2021textile}. However {\note better conclusion}.
% % Problem paragraph
% In particular, the \emph{elasticity} of deformable objects plays a fundamental role for manipulation tasks~\cite{duenser2018interactive}. Modeling the elastic behavior of fabrics has been an important research direction in the field of textile engineering and computer graphics. Both sides managed to provide accurate and realistic solutions by means of complex analytical methods,  but they unfortunately provide little support for robotic applications~\cite{sun2021robotdrlsim}. Another challenge posed from constitutive models is the requirement of the prior knowledge of all the parameters of the objects. Standard approaches to obtain the fabrics mechanical parameters employ highly complex industrial settings that are unsuitable to be used on unconstrained settings~\cite{kawabata1989fabric, minazio1995fast}. To overcome this limitation, recent work has proposed the use of force-feedback data provided by directly interacting with textiles to extract information related to the elastic properties of such objects, allowing their effective identification~\cite{longhini2021textile}.

Our previous work introduced a taxonomy considering the yarn material and the fabric construction
%We have previously demonstrated how a pulling action and force-torque measurements can be used to identify the yarn material and textile construction
technique to understand textile properties~\cite{longhini2021textile}. To meet this goal, it relied on physical interactions (pulling and twisting) and  force-torque measurements.
We build upon this work and encode elastic behaviors of textiles through Elastic Context (EC). To formulate the \ac{ec} we draw inspiration from industrial techniques to characterize textiles, namely the stress-strain curve~\cite{kawabata1989fabric}. These curves can be obtained by recording the forces perceived by a robot when manipulating different textiles. Additionally, we introduce the use of the \ac{ec} to condition \acp{gnn} to model the behaviors of textiles with different elastic properties.% in simulation. %Such models allow both the prediction of task-specific graph dynamics (such as the position of the nodes) and the force-feedback sensed by the robotic end-effectors.

%into a graph representation of a textile sample realised by a graph neural network. This enables learning of  dynamical behaviour of different textile items. The model is trained to predict task specific graph dynamics (position of graph nodes) as well as the force-feedback sensed on the wrists of the robotic end-effector.

%Motivated by such result, in this work we develop a data-driven model of textile elastic dynamics, which leverages our proposed way of encoding of elasticity form robotic sensors named \emph{Elastic Context} (EC).

%We frame the learning problem as a graph dynamics prediction allowing the prediction of both the force-feedback sensed by the robotic grippers as well as the position of the nodes in the textile mesh. 

% Evaluation paragraph
We experimentally evaluate the \ac{ec} and the conditioned GNN on synthetic data obtained in the PyBullet simulator~\cite{coumans2021}, demonstrating the ability of our approach  to encode the complex elastic behavior of a  wide range of textile.
%We extract EC by sampling the specific stress-strain curve of the textiles in $n_{\text{EC}}$ regular intervals. We evaluate the effect of different values of $n_{\text{EC}}$ in simulation and show how our model is able to accurately describe a wide range of textile samples with different elasticities. 
Furthermore, with force measurements collected  on two Franka-Emika Panda robots interacting with textiles, we show 
 that increasing the dimensionality of the \ac{ec} plays an important role in accurately modeling the complex behaviors of textiles when an interactive scenario is considered. Finally, we briefly discuss the limitations of current hardware setups for sensing textile physical properties that go beyond elasticity.
%the limitations of current robotic simulators in modelling the non-linear behavior of textiles by comparing the results of the simulated environment with those of a real-world experiment, with two Franka-Emika Panda robots. We show that considering higher values of $n_{\text{EC}}$ plays an important role in accurately modeling the dynamics of textile objects in the real world. 
In summary, our contributions are:

%Finally, we highlight the challenges of employing EC-models trained in simulation in real-world scenarios considering a sim-to-real force-prediction task and show that the elastic range in real world is more complex than what is currently possible to generate in simulation. In summary:

%%Initially, we evaluate the effect of different ECs to predict the dynamics of textile objects and show the fundamental role of learning representations with EC in a force prediction task with simulated data, highlighting the inability of the standard GNN (without EC) to generalize to different textile samples. 

\begin{itemize}
    \item \emph{Elastic Context}: a way of encoding elastic properties of textiles suitable for robotic manipulation tasks;
    \item A detailed experimental evaluation demonstrating, both in simulation and the real world, that \ac{ec} outperforms methods that do not consider elastic properties when modeling linear and non-linear elastic behaviors of textile. %highlighting the fundamental role of considering EC for force-prediction tasks and modeling both linear and non-linear dynamics of textile objects. 
\end{itemize}
\section{Background}
\label{sec:preliminaries}

%In this section, we provide background information regarding the physical properties of textiles and their role in the robotic manipulation of deformable objects {\comment where in particular we focus on elasticity}.
Physical properties of textiles like elasticity, surface friction, and flexibility are determined by the yarn material and the construction technique~\cite{grishanov2011structure}.  %In this work, we focus on elasticity and we leverage knowledge from textile community to understand the role elasticity plays in robotic manipulation of textiles.
Common yarn materials are cotton, wool, or polyester. In modern textiles, these raw materials are blended with elastomers, such as elastane, making the yarn more flexible~\cite{uyanik2019strength}. The construction technique (woven or knitted) determines how the yarn threads are interlaced or interloped; see our taxonomy in~\cite{longhini2021textile}. Woven textiles, such as jeans, are produced with two sets of tightly interlaced yarns that make the textile rather rigid. Knitted textiles instead, are made with one interloped thread and loops that generate space between parallel running threads, resulting in a stretchable textile. 
These properties play a fundamental role on the interaction dynamics involved in the robotic manipulation of textiles, which can be characterized by the deformation of the textile when an external force, also called stress, is applied~\cite{arriola2020modeling}. 

\begin{figure}
    \centering
    \includegraphics[width=\linewidth]{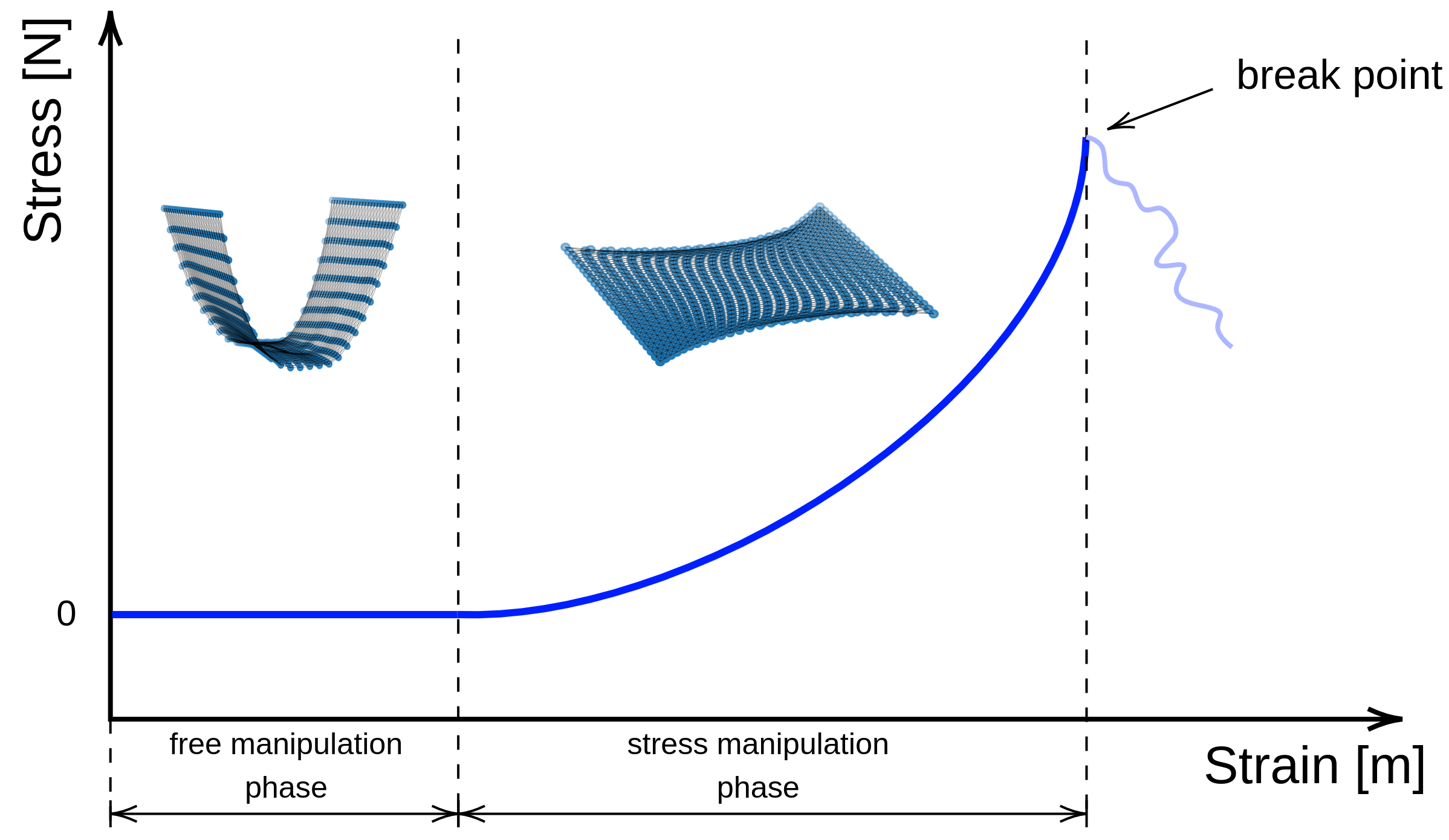}
    \caption{Typical stress curve for manipulation of elastic textiles, categorised in three stages: \emph{free} manipulation, \emph{stress} manipulation and break point. More details in text.}
    \label{fig:phases}
    \vspace{-\baselineskip}
\end{figure}

In particular, these interactions might come either from robot actions, such as stretching or shearing the object, or they could be due to collisions with external objects, like the body of a person in an assistive dressing scenario. In both cases, we can reason about the interaction dynamics by identifying three different stages as shown in Fig.~\ref{fig:phases}. In the first stage (\textit{free} manipulation), no stress is involved during the manipulation as the deformable nature of the textile makes it compliant with the robot's movement or external objects. In the second phase (\textit{stress} manipulation), the textile starts to induce forces on the end-effector and its elastic properties become relevant to characterize the interaction dynamics of the manipulation. Finally, after a certain force known as rupture force, the textile breaks. Often in robotic tasks, objects are manipulated within the \textit{free} manipulation phase. Nevertheless, precise modeling of the \textit{stress} manipulation phase can be useful for applications requiring a constraint over the forces exerted on the deformable object or to assure that a specific action is correctly performed~\cite{erickson2018deep,chi2022iterative}. 
So far, these tasks have been studied with the implicit assumption that the elasticity of different manipulated samples does not exhibit large variations, despite the wide diversity of real-world elastic textiles, such as first-aid bandages, t-shirts, jeans.

The complexity of the tasks robots will be faced with in domestic and industrial setups that consider textiles will require the ability to identify elastic properties online.
%The ability to identify elastic properties online will be crucial for robots to handle the complexity of tasks in both domestic and industrial settings involving textiles.
In textile engineering, the evaluation of the elastic behavior of textiles is performed through the stress-strain test, where a fixed load is applied to a piece of textile and the percentage of displacement, also called strain, is measured~\cite{kawabata1989fabric, eberhardt1996fast}. In robotics, however, we often do not have robots endowed with advanced sensing tools and most of the interaction has historically relied on visual sensing \cite{kragic2002visually}. We need therefore to rely on common force-torque sensors to define a suitable alternative to the stress-strain test. In this work we propose the \ac{ec}, a way of encoding elastic behavior of textiles that can be obtained from the classical robot sensing and can be leveraged by a data driven model to generalize its performances on a wide range of textiles with different elastic properties. 

\section{Related Work}

In this section, we review related work regarding elasticity of textiles from both the analytical and data-driven perspectives. Recent reviews provide an overview of deformable objects, including textiles, and their use in dexterous robotic manipulation~\cite{hou2019review, yin2021modeling}.

\subsection{Analytical Models of Elasticity in Textiles}

Analytical models of textiles usually rely on physics-based modeling and geometric representations, such as particles or graphs~\cite{yin2021modeling}. 
\acp{pbd} approaches model the displacements of a discrete system of particles by applying geometrical constraints~\cite{arriola2020modeling}. This strategy is computationally efficient but struggles to model net forces and is resolution dependent~\cite{eberhardt1996fast}. Graph models overcome these problems by encoding the interaction between the particles in the edges of the graph. In mass-spring models, the edges represent springs following Hooke's law to encode linear elastic relationship \cite{wang2011data}. Damping, bending, and shearing factors can be integrated to reproduce more complex behaviors. However, linear approximations of the textile dynamics underperform with large deformations. The high-strain regime cannot be ignored when modeling textiles as such deformations may occur near the garment’s seams or when parts of the textile are physically constrained. Non-linear models, such as the Neo Hookean or the St. Venant–Kirchhoff, have the advantage of being more accurate than the mass-spring models, but their computational complexity is not suited for real-time simulations~\cite{boonvisut2012estimation,clyde2017modeling, miguel2016modeling}.
%To deal with large displacements, more complex constitutive approaches such as the Neo Hookean or the St. Venant–Kirchhoff models are used to encode non linear hyperelastic relations \cite{boonvisut2012estimation,clyde2017modeling, miguel2016modeling}.  These models have the advantage of being more realistic than the mass-spring models, but their computational complexity is not suitable for fast simulations. 
In robotic manipulation tasks, task-specific and low-dimensional textile representations are often favored over a precise description of all possible mechanical behaviors of the textiles~\cite{mcconachie2020interleaving}. Finding a suitable trade-off between the model accuracy and numerical efficiency is therefore of fundamental importance~\cite{marinkovic2019survey}.

\subsection{Data-driven Models of Textiles}
A limitation of analytical models is the requirement of having precise knowledge of the physical properties of the textile, as these properties are often unknown in real-world scenarios. Despite the effort to simplify the complex measurement tools used in textile engineering, the estimation of textile properties remains highly engineered~\cite{wang2011data, miguel2012data}. Black-box models like neural networks provide a viable solution to model dynamic behavior without explicit knowledge of all the properties. \acp{gnn} have shown to be a suitable framework in the context of learning graph-based dynamics, being employed in a variety of complex domains including deformable objects~\cite{georgousis2021graph, chang2016compositional, lin2022learning}. 
Battaglia,~\etal~\cite{battaglia2016interaction} proposed Interaction Networks (INs) to learn a physical engine to capture local interactions among nodes by modeling them through a complete graph. Li ~\etal~\cite{li2019propagation} extended this framework by proposing Propagation Networks (PNs), which enable instantaneous propagation of forces by multi-step message passing. These approaches have shown excellent results in modeling deformable objects, but they assume the properties of the object to be known a priori. In this work we relax this assumption suggesting to learn a large variety of elastic behaviors by conditioning a GNN on our proposed \emph{Elastic Context}. 

%\input{includes/01_preliminaries}
%\section{Data-Driven Elastic behaviors}

\section{Elastic Context for Data-Driven Models}
\label{sec:method}

To account for the wide range of possible elastic responses that textiles may have, we learn a graph-based dynamics prediction model  using a GNN, which we condition on information about the textile elasticity -  \emph{Elastic Context} (EC). 

\subsection{Elastic Context}
A potential measure of a material's elasticity is the elastic modulus, which evaluates the material's resistance to deformation under stress \cite{arriola2020modeling}. Its value can be derived by measuring the slope of the stress-strain curve corresponding to a specific material or textile sample.  The stress  $\sigma$ is defined as the deformation force $F$ [N] acting on the cross-sectional area $A$ [m$^2$] of %an object
the sample, while the strain $\epsilon$ corresponds to the percentage of displacement $\Delta l$ of the sample with respect to its original length $l_0$.  Using these quantities, the elastic modulus is calculated as $e =\sigma / \epsilon$.

\begin{figure}[t]
  \centering
  \includegraphics[width=1\linewidth]{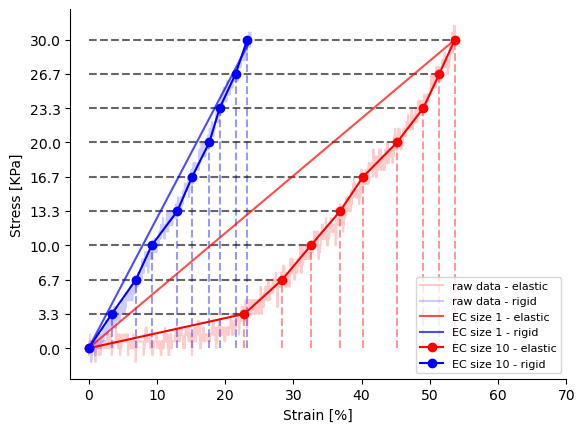}
  \caption{Stress-strain curves of a rigid (blue) and an elastic (red) textile sample:  with $n_{EC}=1$ we recover the elastic modulus, describing a linear elastic behavior that accurately represents the rigid sample (blue) but not the elastic one (red);  with a higher-dimensional EC ($n_{EC}=10$), we are able to better describe the non-linear behavior of the elastic sample.}
  \label{fig:context}
  \vspace{-\baselineskip}
\end{figure}

%An example of stress-strain curves of a rigid and elastic textile samples can be seen in Fig. \ref{fig:context}, where the curves were obtained from force readings  while pulling sample with a dual arm robotic setup (shown in Fig. \ref{fig:envronments}) until a stress $\sigma_{max}$ is reached.

Fig. \ref{fig:context} presents stress-strain curves for two real-world textile samples. The values were obtained from force-feedback readings of the dual-arm robotic setup shown in Fig. \ref{fig:envronments}. The robots were pulling the samples with $l_0 = 0.18$ m and $A=0.18\times 10^{-3}$ m$^2$ until a stress $\sigma_{max}=30$ KPa was reached. 
Fig. \ref{fig:context} indicates how encoding elastic properties of textile with the elastic modulus can only describe small displacements and linear behaviors.
%In fact, the figure shows 
We observe that a linear approximation can accurately describe the rigid sample (blue line) but loses accuracy for the elastic sample (red line), which increases its rigidity with increasing stress.
%he elastic modulus,however, is limited to modeling small displacements and linear behaviors. This is further confirmed by observing in Fig. 3 how a linear approximations well describes the rigid samples (blue line), but loses accuracy for the elastic one (red line) which increases its rigidity with increasing stress

To overcome the limitation of the elastic modulus, we define \ac{ec} as the combination of elastic modules of a given textile evaluated at $n_{EC}$ equidistant points between $0$ and $\sigma_{max}$ on the stress-strain curve, where $n_{EC}$ determines the dimension of the \ac{ec}. We thus represent the \ac{ec} as a vector $EC = [e_1, ..., e_{n_{EC}}] \in \mathbb{R}^{n_{EC}}$, where $e_i = \sigma_i / \epsilon_i$ is the elastic modulus evaluated from the textile stress-strain curve at the corresponding stress $\sigma_i$. Since both stress and strain measurements are normalized by the size of the sample, the \ac{ec} is a consistent definition of elasticity that is comparable between textiles of different sizes as long as $\sigma_{max}$ is fixed.

%{\note this part has to be moved further where a discussion about linear approximation are not always good} 
%The Y-axis represents the force applied per area of the object shape $\sigma = F / A_0$ (the stress), while the X-axis represents the percentage of displacement of the sample over its original length $\epsilon = \Delta L / L_0$  (the strain).
%The Y-axis represents the the stress $\sigma$ defined as the force applied per area of the object shape, while the X-axis represents the strain $\epsilon$ evaluated by observing the percentage of displacement of the sample over its original length. 
%Therefore, the elastic modulus can be obtained as $e =\sigma / \epsilon$. 

\subsection{Graph Model}
We propose to represent textiles as graphs, and to learn their dynamics using \acp{gnn}~\cite{kipf2016semi, sanchez2020learning}. We can thus formulate the problem of learning the force and position dynamics of textiles as learning the parameters $\theta$ of a \ac{gnn} $\Lambda_{\theta}$.
%for a graph dynamics prediction task.
Specifically, let $G_k=(V_k,E_k)$ be a graph representing a textile sample  $k\in K$, where $V$ is the set of nodes, $E$ the set of edges and $K$ is the set of all possible elastic samples. We define the features of each node $v \in V_k$ by their position $x^{t}_{v} \in \mathbb{R}^3$ in the Euclidean space at time step $t$.
The edges  $e \in E_k$ instead describe the elastic relation between two connected nodes. We propose to define the features of the edges as $EC_k \in  \mathbb{R}^{n_{EC}}$ evaluated for the specific elastic textile $k$. $EC$ is therefore a feature shared among all the edges encoding the elastic property of the textile into $G_k$.

At time $t$, a dual arm robotic manipulator grasping the textile applies an action $a^{t} = \Delta x_{v_{grasp}}^t \in \mathbb{R}^3$ at the grasped nodes $v_{grasp} \in V_k$, resulting in new positions $x^{t+1}_{v}$ of the nodes and a textile specific force $F^{t+1}_k$ perceived at the end-effectors. Our goal is to learn a model that leverages the EC to predict  $x^{t+1}_{v}$ and $F^{t+1}_k$, given the initial state of $G_k^{t}$ and the action $a^{t}$ applied to the textile.

%{\note dont like the connection to F, it can be used to state the goal that is to learn to predict it} %$F^{t}_k$ is the force recorded by the force sensors at the two end-effectors of the robot at time $t$, while $a^{t}_{v} = \Delta x_{v} \in \mathbb{R}^3$ is the action applied at the nodes grasped by the robot which updates their features as  $x^{t+1}_{v}= x^{t}_{v} + a^{t}_{v}$.

We employ a standard message-passing architecture for the \ac{gnn} model $\Lambda_{\theta}$  \cite{li2019propagation}. The input graph is constructed by first concatenating the action to the features of the grasped nodes. The features of both nodes and edges are then projected into latent representations, respectively $h_0$ and $c$, through learned encoders parameterized with a \ac{mlp}. Subsequently, every node aggregates messages from its neighbors via $T$ propagation steps, where each propagation leverages the elastic information embedded into the features of the edges to compute the final update of the features of the nodes. In particular, for the propagation step $\tau \in [1, T]$, the features of each node get updated follows:
%Then the graph is projected into a suitable latent representation $h_0$ through an encoder neural network. Subsequently, nodes pass information to their neighbors across T propagation steps. Such propagation considers the information of the edge connecting the nodes and their neighbors $c$, a latent representation of the EC$_k$ encoded through an additional neural network. Both the nodes and edge encoders are parameterized via a \ac{mlp}. At every step $\tau \in [1, T]$ the features of each node get updated following:
\begin{equation}\label{dyna}
h_i^\tau = \Phi\left( \sum_{s\in N_i} \Psi \left( h_i^{\tau-1}, h_s^{\tau-1}, c \right)   \right) \quad \forall v_i \in V_k,
\end{equation}
%\begin{equation}\label{dyna}
% h_0 = \Phi\left( \sum_{s\in N} W_{i,s} \odot \Psi \left( x_i, x_s, c \right)   \right) \forall i\ \in V.
%\end{equation}
where $N_i$ denotes the set of neighbor nodes of node $i$, $\Psi$ is a message-passing network that propagates the information of each node $i$ to its neighbors, and $\Phi$ is an aggregation function of the total information received by each node. We parametrized both $\Psi$ and $\Phi$ via separate MLPs.
Finally, we encode the features of each node $h_i^{T}$ to obtain the estimate of the displacement of each node in $G_k^{t+1}$ and the force-feedback  $F_k^{t+1}$ perceived by the robot in the next time-step $t+1$. This process is done by two projection heads (MLPs), where the final value of $F^{t+1}_k$ is the result of an average pooling layer. 
%Predicting the force $F^{t+1}_k$ along the graph's position can be useful for applications requiring a constraint over the forces exerted on the deformable object, e.g., robotic dressing or bathing assistance {\note this last sentence is more for the introduction rather than the method}.
The overall model $\Lambda$ can be learned using a dataset ${\mathcal{D} = \{(G^t_k, G^{t+1}_k, F^{t+1}_k, a^t_k, EC_k)\}_{\forall k\in K}}$, optimising the parameters $\theta$ using a supervised loss on the prediction of the nodes position and the force exerted at the grasp-nodes:
\begin{equation}
\mathcal{L} =  \mathbb{E}_{\mathcal{D}}\left[ \  d(\Lambda_{\theta}(G^t_k, a^t_k, EC_k), \  (G^{t+1}_k, F^{t+1}_k) \ \right],
\end{equation}
where $d$ is a measure of the distance between the prediction and the ground-truth of graphs and forces. In our case,  $d$ is implemented as the sum of Mean-Squared Error (MSE) of the graph's position and of  the force.

\section{Experimental Evaluation }
In this section, we evaluate the performance of the EC in modeling linear and non-linear dynamics of textile objects for robotic manipulation tasks. In particular, we show using simulation that EC with GNNs leads to more accurate force-feedback predictions of unseen elastic textiles. Furthermore, we analyze the role of the dimensionality of EC both in simulation and real-world scenarios, highlighting the importance of increasing the EC dimension ($n_{EC}$) in presence of non-linear force dynamics. %Finally, we discuss the challenges and limitations of real-world robotic setups to perceive other physical properties of textiles.

\begin{figure}[t]
  \centering
  \includegraphics[width=\columnwidth]{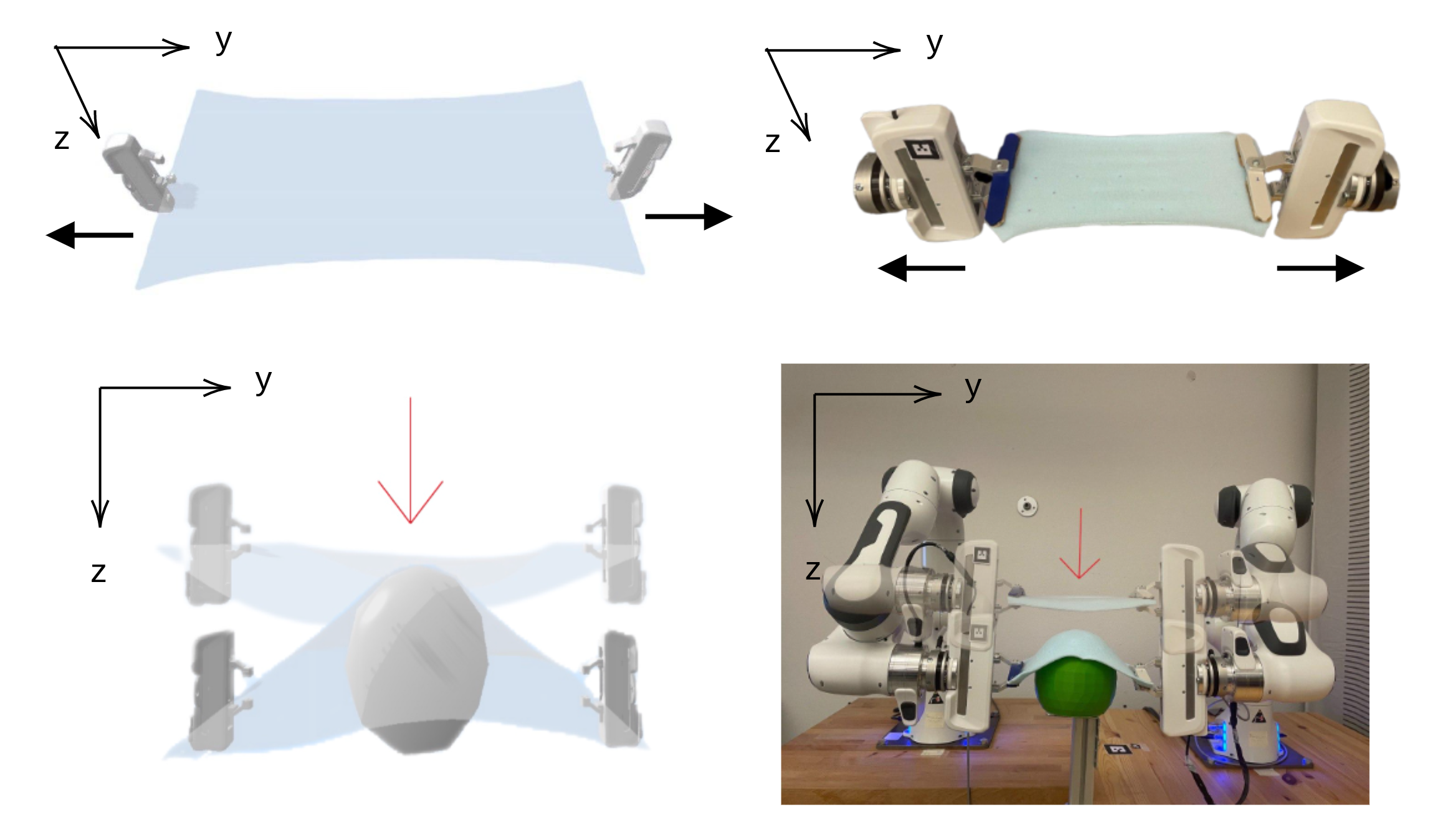}
  \caption{Simulated (left) and real-world (right) environments to instantiate the simplified assistive dressing task: the  \textit{y}-axis corresponds to the pulling direction used to collect the context, while the \textit{z}-axis is related to the task execution.}
  \label{fig:envronments}
  \vspace{-\baselineskip}
\end{figure}

\subsection{Experimental Setup}

\begin{figure*}[t]
     \centering
     \begin{subfigure}[b]{\columnwidth}
         \centering
         \includegraphics[width=\linewidth]{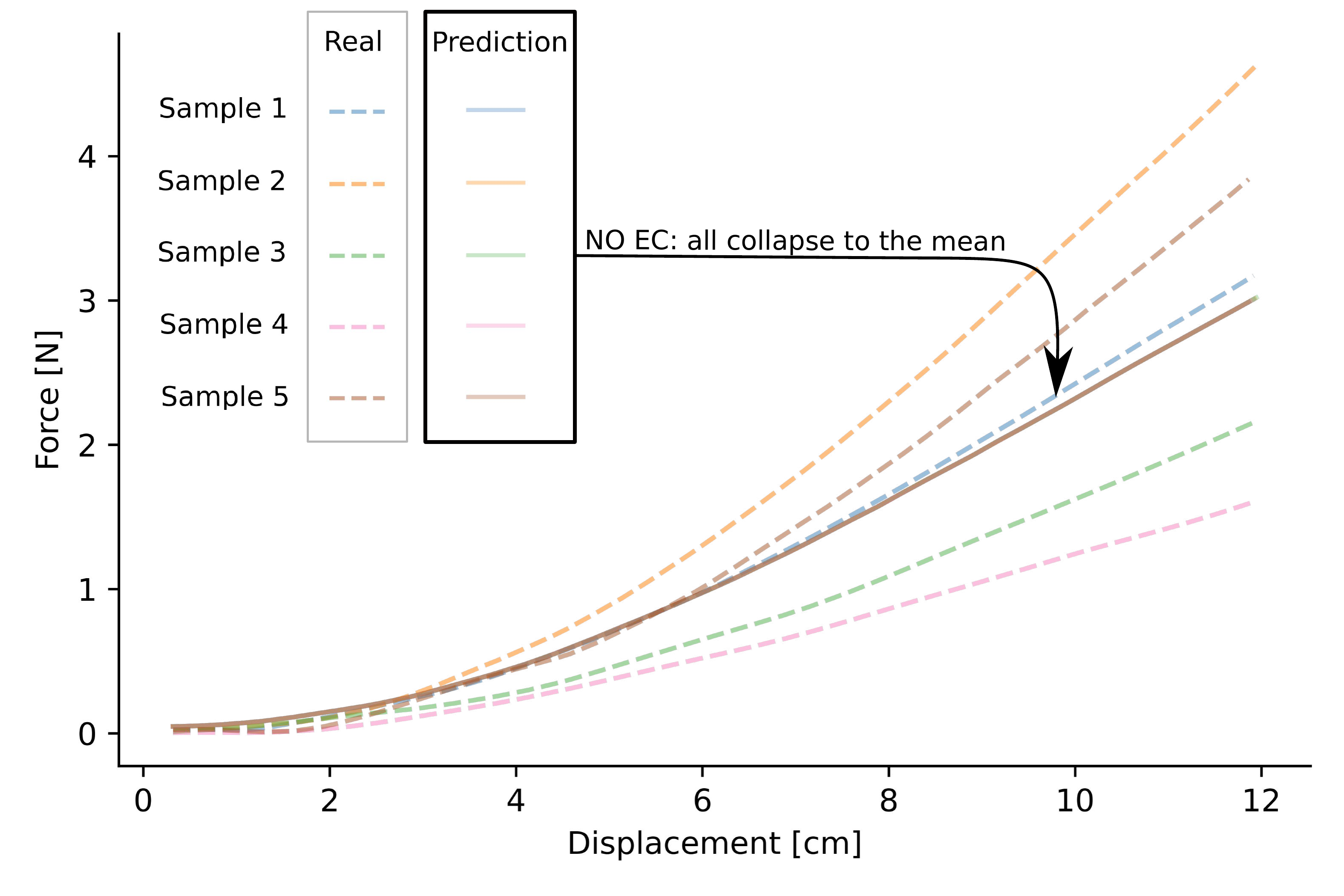}
         \caption{Baseline}
         \label{fig:no_context}
     \end{subfigure}
     \hfill
     \hfill
     \begin{subfigure}[b]{\columnwidth}
         \centering
         \includegraphics[width=\linewidth]{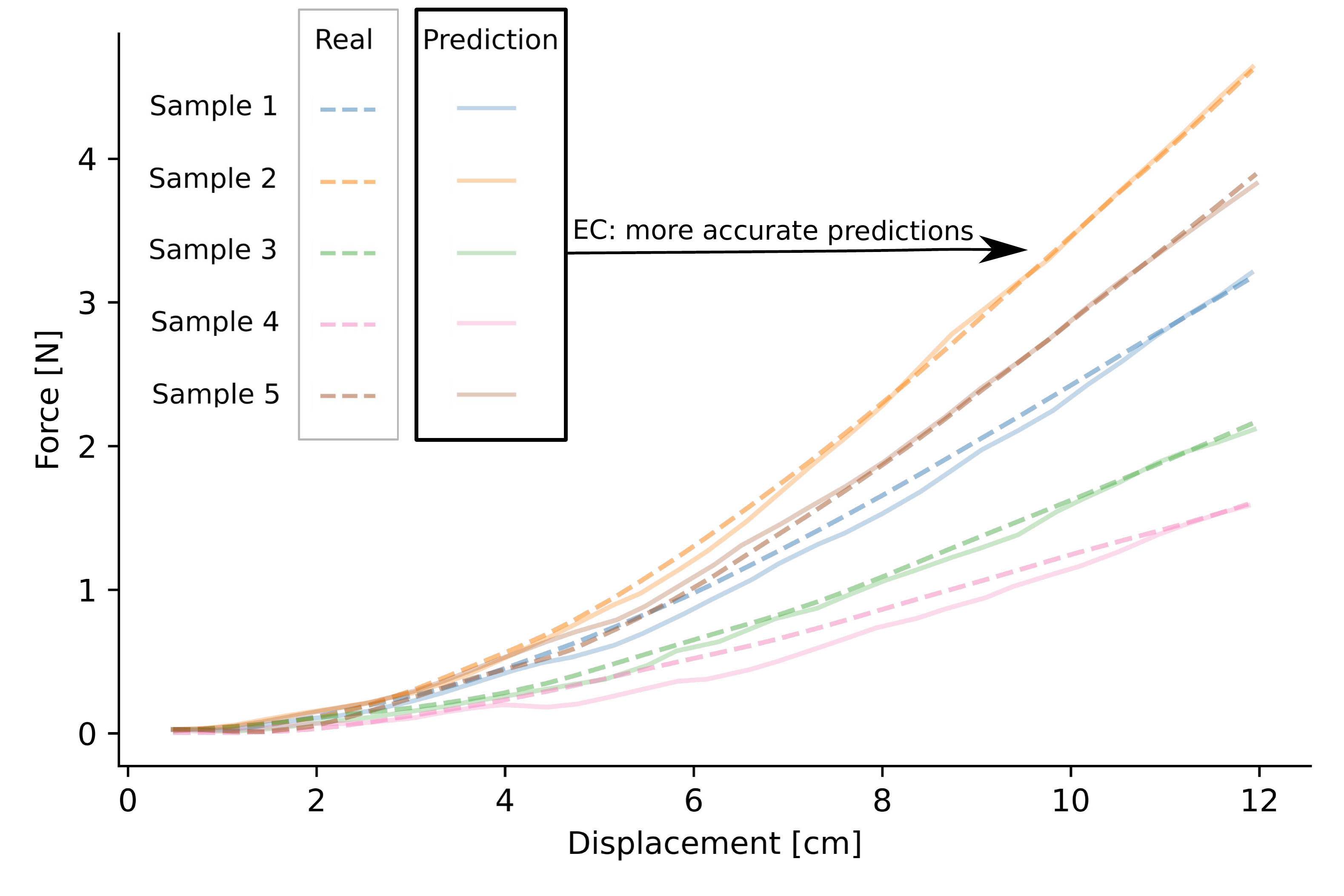}
         \caption{GNN + EC $(n_{EC} = 1)$}
         \label{fig:context_1}
     \end{subfigure}
     \hfill
\caption{Force-forecasting  predictions of the baseline model (a) and the \acp{gnn}  + EC  with dimension $n_{EC} = 1$  model (b) evaluated on 5 test elastic samples. The results show that the  \acs{gnn}  + EC model  generalizes to unseen elastic textiles, leveraging the information provided by the \ac{ec}. }
\label{fig:GNN_context}
\vspace{-\baselineskip}
\end{figure*}

%\subsubsection{Task}
\noindent
\textbf{Task:}
We evaluate the performance of EC in predicting forces perceived by a robot when manipulating \emph{unseen} elastic samples. To this end, we devised a two-stage simplified assistive dressing task. In the first stage, a dual-arm robot is tasked to stretch a textile up to a maximum stress  ${\sigma_{max} = 3\times 10^4}$ Pa, corresponding to a $F_{max} = \sigma_{max} \times A_0$ [N] force perceived at the end-effector for a textile with cross-sectional area $A_0$ [m$^2$]. The forces recorded during this interaction are used to recover the EC, following the procedure defined in Section \ref{sec:method}. In the second stage, the same dual-arm robot is tasked with pulling the sample over a sphere, resembling the head of a person, up to a cumulative gripper displacement of $a_{max}$ [cm] from the instant the textile starts to induce force on the end-effector. 
  
The goal of our model is to perform a \textit{force-forecasting} task, which consists of predicting the force response of the textile from the moment of contact with the object until the goal displacement is reached. 
%The first stretching is needed to recover the EC of the textile sample from the force recordings as described in section \ref{sec:method}. The second pulling instead is the situation where we want our model to perform a \textit{force-forecasting} task, which consists of predicting the force profile for the elastic sample from the moment of contact with the object until the goal displacement is reached. 
We reproduced the aforementioned scenario both in the real world and in the PyBullet simulator~\cite{coumans2021, erickson2020assistive}. An overview of the setup is presented in Fig.~\ref{fig:envronments}, where two robotic grippers either stretch a piece of textile (to obtain the EC) or pull it over a rigid sphere (simplified assistive dressing scenario). To manipulate the textile in the real-world setting we used two Franka-Emika Panda equipped with Optoforce Force/Torque sensors, while in the simulation we only used free-floating end-effectors and their force sensors.

%after a predetermined number of timesteps from the moment of the contact with the object. 
%With this scenario the goal is to highlight the relevance of the EC showing the capability of our model to predict the evolution of the forces of different elastic samples and also analyse the sim-to-real gap {\note this can be put at the beginning as introduction to the task}.  

%\subsubsection{Data Collection}
%\label{sec:datasets}
\noindent
\textbf{Data Collection:}
The simulation dataset ${\mathcal{D}_{SIM} = \{(G^t_k, G^{t+1}_k, F^{t+1}_k, a^t_k, EC_k)\}_{k\in K}}$ was collected performing the aforementioned task with %textile
parameters $A_0 = 0.18 \times 10^{-3}$ m$^2$ and $l_0 = 0.18$ cm. For each execution, we varied the elastic properties $k \in K$ of the simulated textile. $K$ = [20, 119] was defined empirically by selecting the \emph{elasticity} object parameter to avoid unstable behaviors of the mesh during the collision with the sphere. We uniformly sampled $k$ %$k\in K$ 
with a step size of 1 while keeping the \textit{bending} and \textit{damping} properties fixed to $0.1$ and $1.5$ respectively, obtaining a total of $100$ different elastic samples.

The ground truth forces used to obtain EC$_k$ and $F_k^{t+1}$ were recorded from the force sensors on the virtual grippers. We smoothed the force measurements using a Savitzky–Golay filter\cite{savitzky1964smoothing} with a window size of $21$ and a third-grade polynomial to account for noisy measurements due to collision interactions. We obtained $G_k^t$ and $G_k^{t+1}$ by accessing %, from the simulator, 
the ground truth positions of all textile vertices disposed as a $25\times25$ 3D mesh, which we subsequently downsampled to $12\times12$. The gripper actions $a_k^t$ are sequences of 33 displacements along the z-axis in the interval of $[0, a_{max}] $, where $a_{max} = 12$ cm, providing a total of 3300 data samples.

For the real-world experiments, where the focus is on analysing the role of the EC for non-linear force dynamics, we collected a dataset $\mathcal{D}_{RW} = \{(F^{t+1}_k, a^t_k, EC_k)\}_{k\in K}$. We leave the collection of the real-world ground-truth graphs $G^t_k$ and $G^{t+1}_k$ for future work, which could be performed leveraging recent approaches for state estimation and dynamics prediction for cloth ~\cite{lin2022learning, huang2022mesh, EDONET2022}. Differently from the simulated data, the ground truth elastic properties of the textile are not easily accessible as their estimation would require specific tensile tests as discussed in Section~\ref{sec:preliminaries}. To overcome this challenge, we fixed $K$ according to a proxy categorization of textile properties represented by the taxonomy proposed in \cite{longhini2021textile}. We chose $40$ different combinations of yarn material to maximize the variance of the elastic responses, while keeping the construction technique fixed to knitted as the one leading to more elastic behaviors. In particular, we chose the following textile samples classified by their materials: 8 wool, 18 cotton and 10 polyester. Cotton and polyester material classes contain samples with different percentages of elastane.

%accessing the ground truth graph of the textile is a complex task. Therefore, we limited our experimental evaluation of the EC to a dataset $\mathcal{D}_{RW} = \{(F^{t+1}_k, a^t_k, EC_k)\}_{k\in K}$ leaving the collection of $G^t_k$ and $G^{t+1}_k$ for future work. Another problem of the real-world case is the evaluation of the ground truth elastic properties of the textile, as it would require specific tensile tests as discussed in Sec. \ref{sec:preliminaries}. To overcome this challenge, we chose to represent $K$ through a proxy categorization of textile properties defined by the taxonomy proposed in \cite{longhini2021textile}. We chose $40$ different combinations of yarns material and construction technique to maximize the variance of the elastic responses. In particular, we chose {\note X} cotton,{\note X} wool and {\note X}polyester samples blended with different percentages of elastane. {\note $9$} samples are knitted textile, while {\note  $8$} of them are woven.

%\subsubsection{GNN Implementation}
\noindent
\textbf{GNN Implementation:}
To evaluate the role of the EC, we instantiate a GNN with node and edge encoders composed of two linear layers with $16$ neurons each. The message-passing and aggregation functions are implemented as two linear layers with $16$ neurons, and $T=8$. The graph prediction and force prediction heads consist of two linear layers with $16$ hidden neurons, respectively. We used the Rectified Linear Units (ReLU)~\cite{agarap2018deep} as the activation function and layer normalization throughout the network except at the final layers of each block~\cite{ba2016layer}. 
We train the GNN for $2000$ epochs, with a batch size of $32$, and a learning rate of $3\times10^{-4}$. We randomly split the $3300$ data points from simulation with a $0.2$ test-train split, ensuring that the elastic behaviors in the test set are unseen. %different from the ones used for training. %and a test/train split of $0.2$. 
The parameters were optimized using Adam~\cite{kingma2014adam}. Both the training objective and test evaluation metric of the models are the Mean Squared Error (MSE) between the model’s force prediction and the ground truth.

% \begin{figure}[t]
%   \centering
%          \includegraphics[width=0.9\linewidth]{images/rw_exp_3.pdf}
%         %  \caption{Prediction of SIM mesh.}
%     \caption{Example of graph prediction of the GNN + EC ($n_{EC}=1$) evaluated on an test textile sample with input action $a_k^t = 12$ cm. }
%     \label{fig:meshes}
% \end{figure}

\begin{figure*}[t]
     \centering
     \begin{subfigure}[b]{1.\columnwidth}
         \centering
         \includegraphics[width=\linewidth]{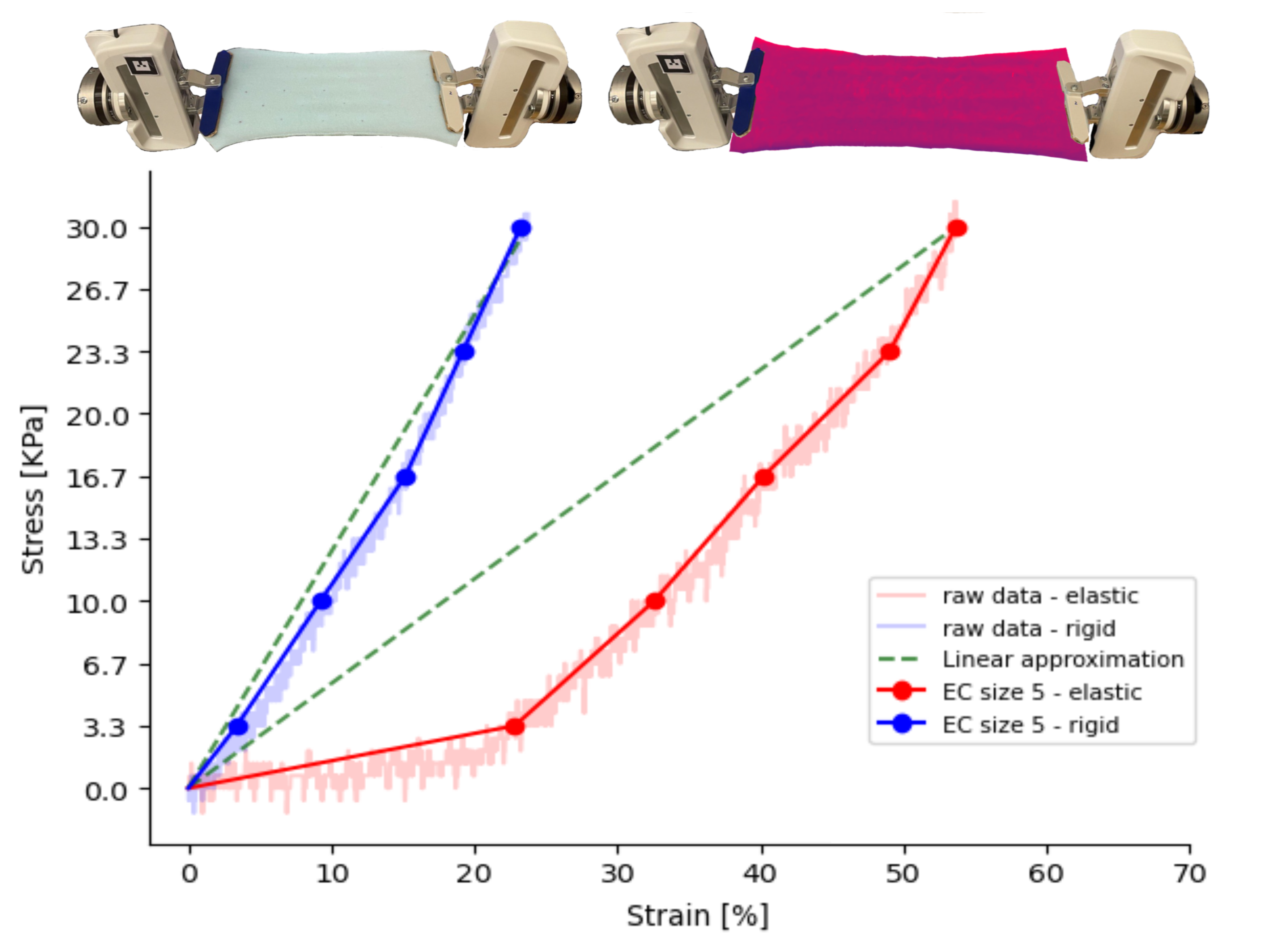}
         \caption{Real-world force profiles}
         \label{fig:rw_profiles}
     \end{subfigure}
     \hfill
     \hfill
     \begin{subfigure}[b]{1.\columnwidth}
         \centering
         \includegraphics[width=\linewidth]{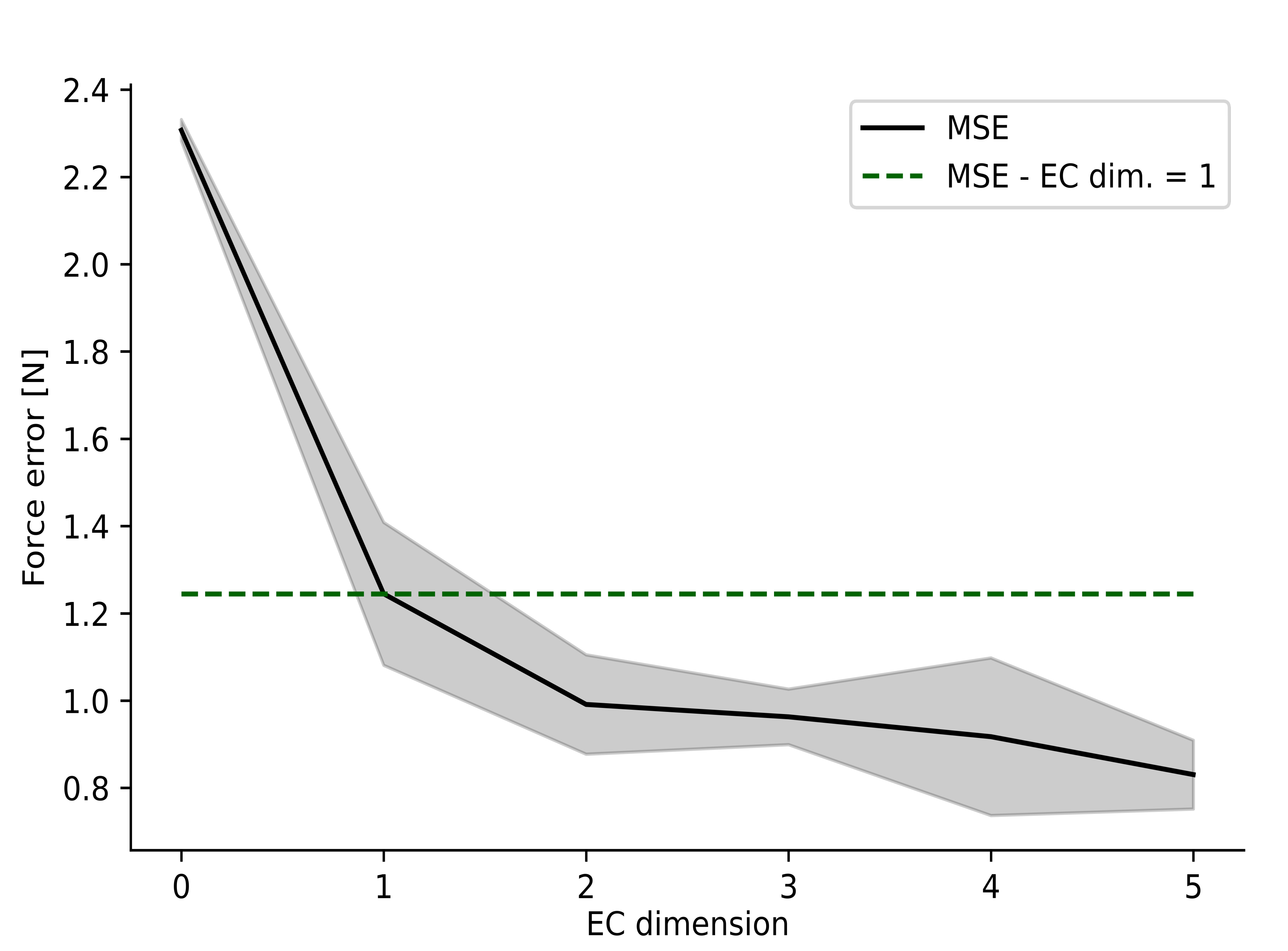}
         \caption{Role of the EC dimension in real world}
         \label{fig:ec_exp}
     \end{subfigure}
     \hfill
\caption{(a) Real world force profiles  of respectively a rigid sample (blue) with an almost linear behavior, and an elastic sample (red) exhibiting a non-linear behavior. (b) \emph{Force forecasting} MSE of the MLP models tested on 8 \emph{unseen} real-world samples.  Results averaged over 6 randomly-seeded runs. }
\vspace{-\baselineskip}
\end{figure*}

\subsection{Elastic Context Evaluation}
We start by evaluating the role of the Elastic Context in the simulation environment, as the ground truth properties of the underlying model of the textile are easily accessible. We carry out the evaluation on the \emph{force-forecasting} task, where the goal is to predict the force evolution and the state (represented as a graph) of unseen elastic textiles up to the goal displacement of $12$ cm. We compare our proposed GNN informed with the EC to a \emph{baseline} GNN that has no %access to the 
EC information, and to an \emph{oracle} GNN that has access to the ground truth properties of the textile. % obtained from the simulator.
Furthermore, we evaluate the effect the dimension of the EC has on simulated force dynamics, where we considered $n_{EC} = \{1, 2, 5\}$. 

Table \ref{tab:GNN_dynamic_force_comparison} presents the test MSE of the force and graph rollout predictions evaluated on 20 \emph{unseen} elastic samples. 
%compared with an MLP baseline sharing the same underlying architecture. 
Regarding the force prediction, the results demonstrate that the baseline model performs worse than the GNNs that have access to elastic information. Moreover, the models that use the EC have comparable performances to the oracle, showing that EC enables the
 models to leverage elastic information. 
These findings are further supported by the qualitative results of the \emph{force forecasting} prediction visualized in Fig. \ref{fig:GNN_context}. The baseline model (Fig,~\ref{fig:no_context}) predicts the same force evolution for all test samples, while the GNN with EC $n_{EC}=1$ (Fig,~\ref{fig:context_1}) 
successfully covers a larger spectrum of the test set, improving the accuracy of the predictions of different elastic behaviors. These results confirm the relevance of encoding elastic behavior of textiles, such as the EC, especially for real-world applications requiring a constraint over the forces exerted on the deformable object
\begin{table}[!h]
    \centering
    \caption{Mean and standard deviation of the MSE of the \emph{force-forecasting} task rollout evaluated on 20 different test samples.}
    \begin{tabular}{lcc}\toprule
         Model & FORCE  error [N]& GRAPH error [m] \\
         
        \toprule
        
         Oracle           &  $0.207 \pm \; \;  0.239$ &  $0.885 \pm  \; \; 0.301$\\ %\midrule
         Baseline         &  $3.169 \pm \; \; 2.434$ &  $0.921 \pm  \; \; 0.292$\\ \toprule
         GNN $n_{EC}=1$   &  $0.269 \pm \; \; 0.301$ &  $0.896 \pm  \; \; 0.234$\\ %\midrule
         GNN $n_{EC}=2$  &  $0.235 \pm \; \; 0.108$ &  $1.108 \pm  \; \; 0.411$\\ %\midrule
         GNN $n_{EC}=5$  &  $0.246 \pm \; \; 0.247$  &  $1.206 \pm  \; \; 0.409$\\ \bottomrule
   
    \end{tabular}
    \label{tab:GNN_dynamic_force_comparison}
\end{table}
(e.g., assistive dressing or bathing assistance). Furthermore, we observe from Table~\ref{tab:GNN_dynamic_force_comparison} that increasing the dimension of the EC does not lead to improvements in the model performance, suggesting that $n_{EC}=1$ is enough to describe the variety of simulated elastic behaviors that assume a linear force dynamics.
% \begin{figure}[h!]
%   \centering
%      \begin{subfigure}[b]{0.45\columnwidth}
%          \centering
%          \includegraphics[width=\linewidth]{images/3D_mesh_pred.jpg}
%          \caption{Prediction of SIM mesh.}
%          \label{fig:sim_mesh}
%      \end{subfigure}
%      \begin{subfigure}[b]{0.45\columnwidth}
%          \centering
%          \includegraphics[width=\linewidth]{images/2D_Mesh_pred.jpg}
%          \caption{Prediction of RW mesh.}
%          \label{fig:RW_mesh}
%      \end{subfigure}
    
%     \vspace{-0.1cm}
%     \caption{ Qualitative results of 3D simulated mesh prediction (left) and 2D real-world mesh prediction(right).}
%     \label{fig:meshes}
% \end{figure}
Regarding  the graph predictions, the quantitative results of the evaluated models are presented in Table~\ref{tab:GNN_dynamic_force_comparison}. It can be noticed that all the models perform comparably, reflecting our design choices of using displacements as actions in combination with textiles with \emph{isotropic} elastic properties. In this scenario, the displacement of the nodes depends solely on the action applied by the robot, which is an information available to all the models. The same consideration does not hold if we consider textiles with anisotropic elastic properties,  which represent an interesting future direction to explore.
%An interesting observation is that there is a consistent improvement of the graph prediction performance for the model with any type of elastic information with respect to the \emph{baseline}. This result indicates that during the interaction with the textile, there are regions of the mesh where the behavior is driven by its elastic properties. In Fig. \ref{fig:meshes} are shown two simulated elastic textiles where the same displacement action is applied. It can be noticed that the edges of the textiles bend differently, suggesting that elastic properties might be observed from positions dynamics as well and could be an interesting future direction to explore.  

\subsection{Real-world Evaluation}

In this section we showcase the relevance of increasing the dimensionality of EC in the presence of non-linear force dynamics. In particular, our goal is to highlight that an EC of dimension 1 represents a good approximation of linear force behaviors, but it loses accuracy for non-linear real-world behaviors as highlighted in Fig.~\ref{fig:rw_profiles}. To fulfill this goal, we implemented an MLP with two hidden layers of 8 units each and ReLU as the activation function. We trained the model with the real-world dataset $\mathcal{D}_{RW}$ to predict $F^{t+1}_k$ given $a^t_k$ and EC$_k$. Moreover, we trained different variants of the model by choosing the dimensionality of the EC in the range ${n_{EC} \in [0, 5]}$, where each of the variants is trained for $1$k epochs. Fig.~\ref{fig:ec_exp} presents the force prediction MSE of each model on 8 test elastic samples averaged over 6 randomly-selected seeds, where the test samples were randomly selected from the dataset for each seed. Contrarily to what was observed for the simulation experiments, these results show that increasing the dimensionality of the EC leads to more accurate predictions of real-world non-linear force profiles with the respect to the models trained with an EC with dimension equal to 1 (dashed line). 
%is relevant to obtain more accurate predictions of real-world non-linear force profiles, lowering the prediction error with the respect to the models trained with an EC with dimension equal to 1 (dashed line).
This outcome highlights the gap between simulated and real-world force dynamics. An interesting future research direction is to explore how this gap hinders the model performance when trained using simulated data and then directly applied to real-world textiles.
We plan to investigate this question in our future work and to address the problem of how to bridge the gap between simulation and real world for better sim-to-real transfer for deformable object manipulation.

%Exploring how this gap between simulated and real-world force dynamics hinders sim2real transfer represents and interest future research direction that we plan to explore.
%Moreover, these results highlight the gap between simulated and real-world force dynamics. An interest future research direction that we plan to explore is how this gap hinders the model performance when trained using simulated data and then directly applied to real-world textiles.

%With this experimental evaluation, we showed how the EC is a viable solution to leverage non-linear elastic behaviors. 

%It is worth noticing that our proposed EC is relevant for describing elastic properties, while factors influencing manipulation of textiles go beyond the non-linear elastic behaviors. Other relevant properties are friction and flexibility, affecting tasks such as ironing or folding.  Furthermore, interesting aspects to study may be the breaking point of textiles or changing properties when textiles are covered with substances like water or oil. All these open new research avenues for the future. 
\section{Conclusions}

In this work, we presented and evaluated \emph{Elastic Context} (EC), an approach to encode elasticity in data-driven models of textiles. We have shown the role of
the EC in a \textit{force forecasting} prediction task on both simulated and real world data. The  EC can be easily employed in real-world robotic platforms, providing a simple way to understand elastic properties of textiles in scenarios where no labels are provided. Such understanding can be of great importance in assistive robotics and human-robot interaction scenarios, where the manipulation of textile and other deformable objects is considered. While the proposed EC is suitable for describing elastic properties, there are other factors that affect textile manipulation beyond non-linear elastic behaviors. For example, friction and flexibility play a role in tasks like ironing and folding. Additionally, studying aspects like the breaking point of textiles or changes in properties when textiles come into contact with substances like water or oil presents new opportunities for future research.
%The proposed EC is suitable for describing elastic properties, while factors influencing manipulation of textiles go beyond the non-linear elastic behaviors, like for example friction and flexibility affecting tasks such as ironing or folding.  Furthermore, interesting aspects to study may be the breaking point of textiles or changing properties when textiles are covered with substances like water or oil. All these open new research avenues for the future. 

\section{Acknowledgements}
This work has been supported by the European Research Council (ERC-BIRD), Swedish Research Council and Knut and Alice Wallenberg Foundation. This material is based upon work supported by the National Science Foundation under NSF CAREER grant number: IIS-2046491

%\section{Acknowledgements}
%This work has been supported by the European Research Council, Swedish Research Council and Knut and Alice Wallenberg Foundation.

\bibliography{references}

\begin{thebibliography}{10}
\providecommand{\url}[1]{#1}
\csname url@samestyle\endcsname
\providecommand{\newblock}{\relax}
\providecommand{\bibinfo}[2]{#2}
\providecommand{\BIBentrySTDinterwordspacing}{\spaceskip=0pt\relax}
\providecommand{\BIBentryALTinterwordstretchfactor}{4}
\providecommand{\BIBentryALTinterwordspacing}{\spaceskip=\fontdimen2\font plus
\BIBentryALTinterwordstretchfactor\fontdimen3\font minus
  \fontdimen4\font\relax}
\providecommand{\BIBforeignlanguage}[2]{{%
\expandafter\ifx\csname l@#1\endcsname\relax
\typeout{** WARNING: IEEEtran.bst: No hyphenation pattern has been}%
\typeout{** loaded for the language `#1'. Using the pattern for}%
\typeout{** the default language instead.}%
\else
\language=\csname l@#1\endcsname
\fi
#2}}
\providecommand{\BIBdecl}{\relax}
\BIBdecl

\bibitem{peters2018review}
B.~S. Peters, P.~R. Armijo, C.~Krause, S.~A. Choudhury, and D.~Oleynikov,
  ``Review of emerging surgical robotic technology,'' \emph{Surgical
  endoscopy}, vol.~32, no.~4, pp. 1636--1655, 2018.

\bibitem{erickson2018deep}
Z.~Erickson, H.~M. Clever, G.~Turk, C.~K. Liu, and C.~C. Kemp, ``Deep haptic
  model predictive control for robot-assisted dressing,'' in \emph{2018 IEEE
  international conference on robotics and automation (ICRA)}.\hskip 1em plus
  0.5em minus 0.4em\relax IEEE, 2018, pp. 4437--4444.

\bibitem{verleysen2020video}
A.~Verleysen, M.~Biondina, and F.~Wyffels, ``Video dataset of human
  demonstrations of folding clothing for robotic folding,'' \emph{The
  International Journal of Robotics Research}, vol.~39, no.~9, pp. 1031--1036,
  2020.

\bibitem{klee2015personalized}
S.~D. Klee, B.~Q. Ferreira, R.~Silva, J.~P. Costeira, F.~S. Melo, and
  M.~Veloso, ``Personalized assistance for dressing users,'' in
  \emph{International Conference on Social Robotics}.\hskip 1em plus 0.5em
  minus 0.4em\relax Springer, 2015, pp. 359--369.

\bibitem{kapusta2019personalized}
A.~Kapusta, Z.~Erickson, H.~M. Clever, W.~Yu, C.~K. Liu, G.~Turk, and C.~C.
  Kemp, ``Personalized collaborative plans for robot-assisted dressing via
  optimization and simulation,'' \emph{Autonomous Robots}, vol.~43, no.~8, pp.
  2183--2207, 2019.

\bibitem{chi2022iterative}
C.~Chi, B.~Burchfiel, E.~Cousineau, S.~Feng, and S.~Song, ``Iterative residual
  policy: for goal-conditioned dynamic manipulation of deformable objects,''
  \emph{arXiv preprint arXiv:2203.00663}, 2022.

\bibitem{ha2022flingbot}
H.~Ha and S.~Song, ``Flingbot: The unreasonable effectiveness of dynamic
  manipulation for cloth unfolding,'' in \emph{Conference on Robot
  Learning}.\hskip 1em plus 0.5em minus 0.4em\relax PMLR, 2022, pp. 24--33.

\bibitem{poincloux2018geometry}
S.~Poincloux, M.~Adda-Bedia, and F.~Lechenault, ``Geometry and elasticity of a
  knitted fabric,'' \emph{Physical Review X}, vol.~8, no.~2, p. 021075, 2018.

\bibitem{clyde2017modeling}
D.~Clyde, J.~Teran, and R.~Tamstorf, ``Modeling and data-driven parameter
  estimation for woven fabrics,'' in \emph{Proceedings of the ACM
  SIGGRAPH/Eurographics Symposium on Computer Animation}, 2017, pp. 1--11.

\bibitem{sperl2020homogenized}
G.~Sperl, R.~Narain, and C.~Wojtan, ``Homogenized yarn-level cloth.'' \emph{ACM
  Trans. Graph.}, vol.~39, no.~4, p.~48, 2020.

\bibitem{eberhardt1996fast}
B.~Eberhardt, A.~Weber, and W.~Strasser, ``A fast, flexible, particle-system
  model for cloth draping,'' \emph{IEEE Computer Graphics and Applications},
  vol.~16, no.~5, pp. 52--59, 1996.

\bibitem{yousef2018investigating}
M.~I. Yousef and G.~K. Stylios, ``Investigating the challenges of measuring
  combination mechanics in textile fabrics,'' \emph{Textile Research Journal},
  vol.~88, no.~23, pp. 2741--2754, 2018.

\bibitem{longhini2021textile}
A.~Longhini, M.~C. Welle, I.~Mitsioni, and D.~Kragic, ``Textile taxonomy and
  classification using pulling and twisting,'' \emph{arXiv preprint
  arXiv:2103.09555}, 2021.

\bibitem{kawabata1989fabric}
S.~Kawabata and M.~Niwa, ``Fabric performance in clothing and clothing
  manufacture,'' \emph{Journal of the Textile Institute}, vol.~80, no.~1, pp.
  19--50, 1989.

\bibitem{coumans2021}
E.~Coumans and Y.~Bai, ``Pybullet, a python module for physics simulation for
  games, robotics and machine learning,'' \url{http://pybullet.org},
  2016--2021.

\bibitem{grishanov2011structure}
S.~Grishanov, ``Structure and properties of textile materials,'' in
  \emph{Handbook of textile and industrial dyeing}.\hskip 1em plus 0.5em minus
  0.4em\relax Elsevier, 2011, pp. 28--63.

\bibitem{uyanik2019strength}
S.~Uyanik and K.~H. Kaynak, ``Strength, fatigue and bagging properties of
  plated plain knitted fabrics containing different rates of elastane,''
  \emph{International Journal of Clothing Science and Technology}, 2019.

\bibitem{arriola2020modeling}
V.~E. Arriola-Rios, P.~Guler, F.~Ficuciello, D.~Kragic, B.~Siciliano, and J.~L.
  Wyatt, ``Modeling of deformable objects for robotic manipulation: A tutorial
  and review,'' \emph{Frontiers in Robotics and AI}, p.~82, 2020.

\bibitem{kragic2002visually}
D.~Kragi{\'c}, L.~Petersson, and H.~I. Christensen, ``Visually guided
  manipulation tasks,'' \emph{Robotics and Autonomous Systems}, vol.~40, no.
  2-3, pp. 193--203, 2002.

\bibitem{hou2019review}
Y.~C. Hou, K.~S.~M. Sahari, and D.~N.~T. How, ``A review on modeling of
  flexible deformable object for dexterous robotic manipulation,''
  \emph{International Journal of Advanced Robotic Systems}, vol.~16, no.~3, p.
  1729881419848894, 2019.

\bibitem{yin2021modeling}
H.~Yin, A.~Varava, and D.~Kragic, ``Modeling, learning, perception, and control
  methods for deformable object manipulation,'' \emph{Science Robotics},
  vol.~6, no.~54, 2021.

\bibitem{wang2011data}
H.~Wang, J.~F. O'Brien, and R.~Ramamoorthi, ``Data-driven elastic models for
  cloth: modeling and measurement,'' \emph{ACM transactions on graphics (TOG)},
  vol.~30, no.~4, pp. 1--12, 2011.

\bibitem{boonvisut2012estimation}
P.~Boonvisut and M.~C. {\c{C}}avu{\c{s}}o{\u{g}}lu, ``Estimation of soft tissue
  mechanical parameters from robotic manipulation data,'' \emph{IEEE/ASME
  Transactions on Mechatronics}, vol.~18, no.~5, pp. 1602--1611, 2012.

\bibitem{miguel2016modeling}
E.~Miguel, D.~Miraut, and M.~A. Otaduy, ``Modeling and estimation of
  energy-based hyperelastic objects,'' in \emph{Computer Graphics Forum},
  vol.~35, no.~2.\hskip 1em plus 0.5em minus 0.4em\relax Wiley Online Library,
  2016, pp. 385--396.

\bibitem{mcconachie2020interleaving}
D.~McConachie, M.~Ruan, and D.~Berenson, ``Interleaving planning and control
  for deformable object manipulation,'' in \emph{Robotics Research}.\hskip 1em
  plus 0.5em minus 0.4em\relax Springer, 2020, pp. 1019--1036.

\bibitem{marinkovic2019survey}
D.~Marinkovic and M.~Zehn, ``Survey of finite element method-based real-time
  simulations,'' \emph{Applied Sciences}, vol.~9, no.~14, p. 2775, 2019.

\bibitem{miguel2012data}
E.~Miguel, D.~Bradley, B.~Thomaszewski, B.~Bickel, W.~Matusik, M.~A. Otaduy,
  and S.~Marschner, ``Data-driven estimation of cloth simulation models,'' in
  \emph{Computer Graphics Forum}, vol.~31, no. 2pt2.\hskip 1em plus 0.5em minus
  0.4em\relax Wiley Online Library, 2012, pp. 519--528.

\bibitem{georgousis2021graph}
S.~Georgousis, M.~P. Kenning, and X.~Xie, ``Graph deep learning: State of the
  art and challenges,'' \emph{IEEE Access}, 2021.

\bibitem{chang2016compositional}
M.~B. Chang, T.~Ullman, A.~Torralba, and J.~B. Tenenbaum, ``A compositional
  object-based approach to learning physical dynamics,'' \emph{arXiv preprint
  arXiv:1612.00341}, 2016.

\bibitem{lin2022learning}
X.~Lin, Y.~Wang, Z.~Huang, and D.~Held, ``Learning visible connectivity
  dynamics for cloth smoothing,'' in \emph{Conference on Robot Learning}.\hskip
  1em plus 0.5em minus 0.4em\relax PMLR, 2022, pp. 256--266.

\bibitem{battaglia2016interaction}
P.~W. Battaglia, R.~Pascanu, M.~Lai, D.~Rezende, and K.~Kavukcuoglu,
  ``Interaction networks for learning about objects, relations and physics,''
  \emph{arXiv preprint arXiv:1612.00222}, 2016.

\bibitem{li2019propagation}
Y.~Li, J.~Wu, J.-Y. Zhu, J.~B. Tenenbaum, A.~Torralba, and R.~Tedrake,
  ``Propagation networks for model-based control under partial observation,''
  in \emph{2019 International Conference on Robotics and Automation
  (ICRA)}.\hskip 1em plus 0.5em minus 0.4em\relax IEEE, 2019, pp. 1205--1211.

\bibitem{kipf2016semi}
T.~N. Kipf and M.~Welling, ``Semi-supervised classification with graph
  convolutional networks,'' \emph{arXiv preprint arXiv:1609.02907}, 2016.

\bibitem{sanchez2020learning}
A.~Sanchez-Gonzalez, J.~Godwin, T.~Pfaff, R.~Ying, J.~Leskovec, and
  P.~Battaglia, ``Learning to simulate complex physics with graph networks,''
  in \emph{International Conference on Machine Learning}.\hskip 1em plus 0.5em
  minus 0.4em\relax PMLR, 2020, pp. 8459--8468.

\bibitem{erickson2020assistive}
Z.~Erickson, V.~Gangaram, A.~Kapusta, C.~K. Liu, and C.~C. Kemp, ``Assistive
  gym: A physics simulation framework for assistive robotics,'' in \emph{2020
  IEEE International Conference on Robotics and Automation (ICRA)}.\hskip 1em
  plus 0.5em minus 0.4em\relax IEEE, 2020, pp. 10\,169--10\,176.

\bibitem{savitzky1964smoothing}
A.~Savitzky and M.~J. Golay, ``Smoothing and differentiation of data by
  simplified least squares procedures.'' \emph{Analytical chemistry}, vol.~36,
  no.~8, pp. 1627--1639, 1964.

\bibitem{huang2022mesh}
Z.~Huang, X.~Lin, and D.~Held, ``Mesh-based dynamics with occlusion reasoning
  for cloth manipulation,'' \emph{arXiv preprint arXiv:2206.02881}, 2022.

\bibitem{EDONET2022}
\BIBentryALTinterwordspacing
A.~Longhini, M.~Moletta, A.~Reichlin, M.~C. Welle, D.~Held, Z.~Erickson, and
  D.~Kragic, ``Edo-net: Learning elastic properties of deformable objects from
  graph dynamics,'' 2022. [Online]. Available:
  \url{https://arxiv.org/abs/2209.08996}
\BIBentrySTDinterwordspacing

\bibitem{agarap2018deep}
A.~F. Agarap, ``Deep learning using rectified linear units (relu),''
  \emph{arXiv preprint arXiv:1803.08375}, 2018.

\bibitem{ba2016layer}
J.~L. Ba, J.~R. Kiros, and G.~E. Hinton, ``Layer normalization,'' \emph{arXiv
  preprint arXiv:1607.06450}, 2016.

\bibitem{kingma2014adam}
D.~P. Kingma and J.~Ba, ``Adam: A method for stochastic optimization,''
  \emph{arXiv preprint arXiv:1412.6980}, 2014.

\end{thebibliography}
\bibliographystyle{IEEEtran}

\end{document}